\documentclass{article}
\usepackage{spconf,amsmath,graphicx}
\usepackage{color}
\usepackage{amssymb}
\usepackage{times}
\usepackage{epsfig,epstopdf}
\usepackage{graphicx}
\usepackage{subcaption}
\usepackage{stfloats}
\usepackage{float}
\usepackage{amsmath,bm}
\usepackage{amssymb}
\usepackage{float}
\usepackage[table,xcdraw]{xcolor}
 \usepackage{multirow}
\usepackage{hyperref}

\def\sqz{\vspace{-3pt}}
\newcommand{\rd}[1]{\textcolor{red}{#1}}
\newcommand{\bl}[1]{\textcolor{blue}{#1}}

\title{\vspace{-15pt}Orthogonally Regularized Deep Networks \\For Image Super-resolution}
\name{\vspace{-15pt}Tiantong Guo, Hojjat S. Mousavi, Vishal Monga \thanks{This work is supported by an NSF CAREER Award to V. Monga.}}
\address{The Pennsylvania State University}
\begin{document}

%
\maketitle
\begin{abstract}\sqz
Deep learning methods, in particular trained Convolutional Neural Networks (CNNs) have recently been shown to produce compelling state-of-the-art results for single image Super-Resolution (SR). Invariably, a CNN is learned to map the low resolution (LR) image to its corresponding high resolution (HR) version in the spatial domain. Aiming for faster inference and more efficient solutions than solving the SR problem in the spatial domain, we propose a novel network structure for learning the SR mapping function in an image transform domain, specifically the Discrete Cosine Transform (DCT). As a first contribution, we show that DCT can be integrated into the network structure as a Convolutional DCT (CDCT) layer. We further extend the network to allow the CDCT layer to become {\em trainable (i.e. optimizable)}. Because this layer represents an image transform, we enforce pairwise orthogonality constraints on the individual basis functions/filters. This Orthogonally Regularized Deep SR network (ORDSR) simplifies the SR task by taking advantage of image transform domain while adapting the design of transform basis to the training image set. 
Experimental results show ORDSR achieves state-of-the-art SR image quality with fewer parameters than most of the deep CNN methods.
\end{abstract}
%
%
\sqz\sqz
\section{Introduction}\sqz\sqz
\label{sec:intro}

Single Image Super-Resolution (SISR) has emerged as one of the most significant ill-posed imaging problems due to a variety of applications in civilian domains as well as in law enforcement \cite{park2003super}. With an increasing number of mobile cameras, generating a clean, sharp image with lower storage and computation requirements is highly desirable.

The single image SR task has been addressed by dictionary based and sparsity constrained learning methods and more recently via deep learning methods. A typical learning/example based SR approach employs two dictionaries of HR/LR images/patches \cite{mallat2010super,chang2004super,glasner2009super,yang2010image,kim2010single}. These dictionaries are often learned with sparse coding methods to reconstruct the SR results. Many of these methods require handcrafted dictionary features which are not readily available \cite{zhang17image}.

Recently, deep learning methods have shown to produce compelling state-of-the-art SR results and across a variety of different image collections \cite{Timofte_2017_CVPR_Workshops}. One of the earliest deep SR methods was SRCNN \cite{dong2014learning,dong2016accelerating} and its extensions that train multiple coupled networks have been pursued as well \cite{tiantong16deep}. Other variants include \cite{huang2015single,wang2015self} which use self-similar patches to explore the self-example based SR idea. However, the network structures are no-less mutations of straightforward spatial mapping functions between LR/HR image. These spatial domain mappings were further boosted by global and local by-pass structures as introduced by residual learning \cite{Kim_2016_VDSR}. Residual network structures essentially reduce the training burden (in the sense of learning complexity) of the deep CNN which is still constructed in spatial domain.  

\textbf{Motivation}: Our work is motivated by the recent promising performance of SR methods in the transform domain \cite{Timofte_2017_CVPR_Workshops}. Our goal is faster inference and structures with fewer parameters than existing spatial domain CNNs. Specifically, the Discrete Wavelet Transformation (DWT) has been explored for the SR problem in traditional frameworks \cite{zhao2003wavelet,robinson2010efficient,wahed2007image,ji2009robust} and more recently also in deep networks \cite{guo2017deep}.  

In this paper, we begin by exploring a DCT domain deep SR method. In the DCT domain, the differences between a given LR-HR image pair is the missing high-frequency information while they typically share the same low-frequency signature (see analysis in section \ref{sec:format}). Because the low-to-high-resolution mapping is simpler, the learning burden of the network can be reduced and both the convergence rate and inference of the network can become faster.  As a first contribution, we show that DCT can be integrated into the network structure as a convolutional DCT (CDCT) layer. We further extend the network to allow the CDCT layer to become {\em trainable (i.e. optimizable)}. Because this layer represents an image transform, we enforce pairwise orthogonality constraints on the individual basis functions/filters. This Orthogonally Regularized Deep SR network (ORDSR) simplifies the SR task by taking advantage of image transform domain while adapting the design of transform basis to the training image set. 


The main contributions of this paper are as follows:
\begin{enumerate} \setlength{\parskip}{-1pt} \sqz
	\item We propose a novel network structure that attacks SR problem in the image transform domain;\sqz
	\item We build a special CDCT layer integrating DCT procedure into the network, where the CDCT filters are adaptable and trainable;\sqz
	\item We add novel orthogonality constraints on the newly introduced `transform layer' to maintain the pairwise orthogonality properties of the learned basis.\sqz
\end{enumerate}
To the best of our knowledge, ORDSR network is the first approach that allows optimization of basis functions for transform domain image SR within a deep learning framework. 

\section{Super-Resolution in DCT domain}\sqz\sqz
\label{sec:format}

An image $x(n_1, n_2)$ of size $H\times W$ can be decomposed into $H/N\times W/N$ blocks of size $N\times N$. For the $(m,n)^{th}$ block, the DCT coefficients are computed as:\sqz\sqz
\begin{equation}\resizebox{0.9\linewidth}{!}{$
X_{m,n}(k_1,k_2) = \sum\limits_{n_2=0}^{N-1}\sum\limits_{n_1=0}^{N-1} x_{m,n}(n_1,n_2)\times w_{k_1,k_2}(n_1,n_2)$}
\end{equation}
where $k_1, k_2, n_1, n_2 = 0,\ldots, N-1$, and $w_{k_1, k_2}(n_1, n_2)$ is the DCT basis function, specifically DCT-II basis, defined as:\sqz
\begin{equation}\resizebox{0.9\linewidth}{!}{$
w_{k_1, k_2}(n_1, n_2) = C_{k_1,k_2}cos\left[\frac{\pi}{N}\left(n_1+\frac{1}{2}\right)k_1\right]cos\left[\frac{\pi}{N}\left(n_2+\frac{1}{2}\right)k_2\right]$}
\end{equation}
where $C_{k_1, k_2} = \frac{\sqrt{1+\delta_{k_1}}\sqrt{1+\delta_{k_2}}}{N}$ and $\delta_k=1$ if $k=0$, $\delta_k=0$ otherwise. For $N=8$, there are $8\times 8$ DCT bases and each basis $w_{k_1,k_2}$ is of size $8\times 8$.

Basis functions $\{w_{k_1,k_2}\}_{k_1,k_2 =1,1}^{N,N}\in\mathbb{R}^{N\times N}$ are pairwise orthogonal, forming an orthogonal basis family:\sqz\sqz
\begin{equation}
<w_{k_1,k_2},w_{l_1,l_2}>=\begin{cases}1,&\text{if }k_1=l_1, \text{ and }k_2=l_2\\0,&\text{Otherwise}\end{cases} \label{equ:orthoDCT}
\end{equation}

Corresponding to the DCT, the inverse DCT (IDCT) for the $(m,n)^{th}$ block is computed as:\sqz
\begin{equation}\resizebox{0.9\linewidth}{!}{$
x_{m,n}(n_1,n_2) = \sum\limits_{k_2=0}^{N-1}\sum\limits_{k_1=0}^{N-1}X_{m,n}(k_1,k_2)\times w_{k_1,k_2}(n_1,n_2)$}
\end{equation}
Note that classical DCT is performed on $N\times N$ blocks of the original image. We now develop a reorganization of the DCT coefficients and their computation, which we show in Section \ref{sec:dctLayer} helps facilitate the implementation of DCT within a CNN.

\begin{figure}
		\centering
		\includegraphics[trim=0cm 0.2cm 0cm 0cm,clip,width=0.9\linewidth]{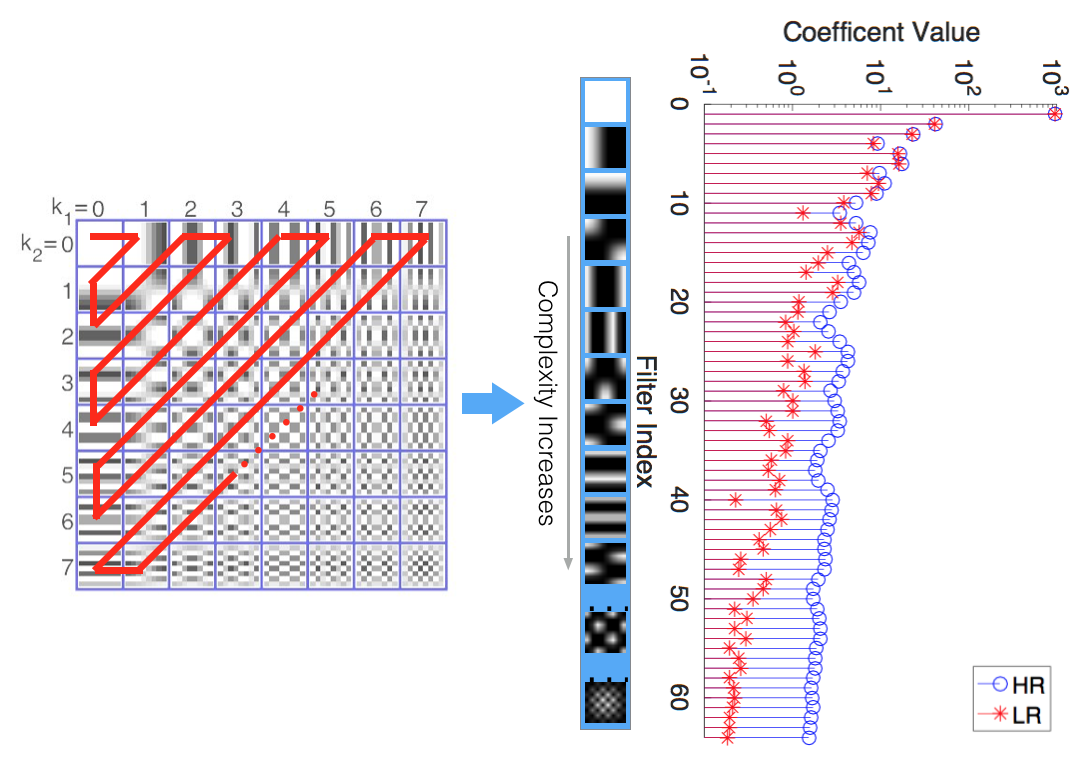}\sqz\sqz\sqz
    \caption{\ninept \textit{Left}: zig-zag reorder of the DCT basis family. \textit{Right}: average coefficient values generated by $\{w_i\}_{i= 1}^{64}$ of \textit{lenna.bmp}.}\label{fig:zigzag}\sqz\sqz\sqz\sqz\sqz
		\end{figure}

\textbf{Zig-zag reorder}: We treat DCT basis functions as filters and reorganize them in a zig-zag order as shown in Fig. \ref{fig:zigzag}.
The zig-zag function maps $\{w_{k_1,k_2}\}_{k_1,k_2=1,1}^{N, N}$ to $\{w_i\}_{i= 1}^{N\times N}$. Specially, with the zig-zag mapping, as the index $i$ increases, the complexity of $w_i$ also increases, \textit{i.e.} the lower end of $\{w_i\}_{i= 1}^{64}$ is corresponding to low-frequency filters, while the higher end (bigger $i$) represents the high-frequency ones.

    Given an HR image $y$, and its LR version $x$, we can plot the average coefficient values generated by the DCT filters $\{w_i\}_{i= 1}^{N\times N}$, as shown in Fig. \ref{fig:zigzag}.
    As the plot suggests, the HR image $y$ and the LR image $x$ share the same low-frequency spectra, while $x$ has less high-frequency information than $y$. With the help of DCT filters, SR becomes a problem of recovering high-frequency DCT coefficients of the HR image from the corresponding ones of the LR input.
\section{Convolutional DCT layer with the orthogonality constraints} \sqz\sqz
\label{sec:dctLayer}

To integrate the DCT analysis within a CNN, we construct a convolutional DCT (CDCT) layer. 

\textbf{Initialization}: The CDCT layer is initialed using the DCT basis $\{w_i\}_{i= 1}^{N\times N}$. For $N=8$, there are $64$ filters $\{w_i\}_{i= 1}^{64}$ of size $8\times 8$ in the CDCT layer such that the complexity (high-frequency content) increases with the filter index.

Unlike classical DCT that produces $8\times 8$ block-wise DCT coefficients, the CDCT layer produces $64$ frequency maps $\{f_i\}_{i= 1}^{64}$ for the whole image by convolving $\{w_i\}_{i= 1}^{64}$ with the input image $x$ as shown in Eq. (\ref{equ:CDCTconv}).\sqz\sqz
\begin{equation}
f_i = w_i * x \label{equ:CDCTconv}, \forall i \in [1,...,64]\sqz\sqz
\end{equation}
These maps, $\{f_i\}_{i= 1}^{64}$, form a cube called DCT cube. The DCT cube is essentially a reorganized version of classical block-wise DCT coefficients of the whole image. 

As $i$ increases, $f_i$ corresponds to higher frequency components of the whole image. Thus, we divide the DCT cube into two parts by a threshold $T$, namely low-frequency spectral maps $f_{low} = \{f_i\}^T_{i=1}$ and high-frequency spectral maps $f_{high} = \{f_i\}^{64}_{i=T +1}$.

The CDCT layer can also perform IDCT by  transpose convolving\footnote{Some literature \cite{noh2015learning,dumoulin2016guide} refer this procedure as deconvolution,  fractionally stride convolution or backward convolution in neural network setups.} $\{w_i\}_{i= 1}^{64}$ with the DCT cube $\{f_i\}_{i= 1}^{64}$ respectively, resulting in the spatial image $y$. This procedure can be viewed as a convolution of $w_i$ with a 8-zero padded $f_i$:\sqz\sqz\sqz
\begin{equation}\sqz\sqz
y = \sum_{i=1}^{64} w_i * g(f_i) \label{equ:CDCTdeconv}
\end{equation}
where $g(\cdot)$ is the zero padding function. For details on the implementation of both the DCT and IDCT as a CNN layer we refer the reader to our accompanying technical report \cite{techReport}.

\textbf{Orthogonality Constraints}: The aforementioned CDCT layer can in fact be learned. Consistent with classical DCT, we enable learning but in the presence of pairwise orthogonality constraints. These constraints are captured by a regularization term which is added to the network's total cost function -- see Eq. (\ref{eq:finalCostFn}). As suggested in (\ref{equ:orthoDCT}), any distinct filter pairs in the CDCT layer should have a zero inner product.
Here, the inner product is computed by vectorized multiplication between two filters. Ideally, $\epsilon$ should be zero but may be relaxed slightly in practice for numerical optimization.\sqz\sqz
\begin{equation}
\forall i\neq j, \|vec(w_i)^Tvec(w_j)-\epsilon\|^2_2=0\label{equ:orthoCDCT}
\end{equation}

\section{ORDSR Network Structure}\sqz\sqz
\label{sec:ORD}
The ORDSR network has two parts: a CDCT layer and a $D-$layer CNN. 
The CDCT layer produces both the DCT cube of the input image and generates the SR results from the CNN's output DCT cube. The CNN recovers the high-frequency spectra by generating a SR-DCT cube.

\begin{figure}
\centering
\includegraphics[trim=1cm 0cm 0cm 0cm,clip,width=\linewidth]{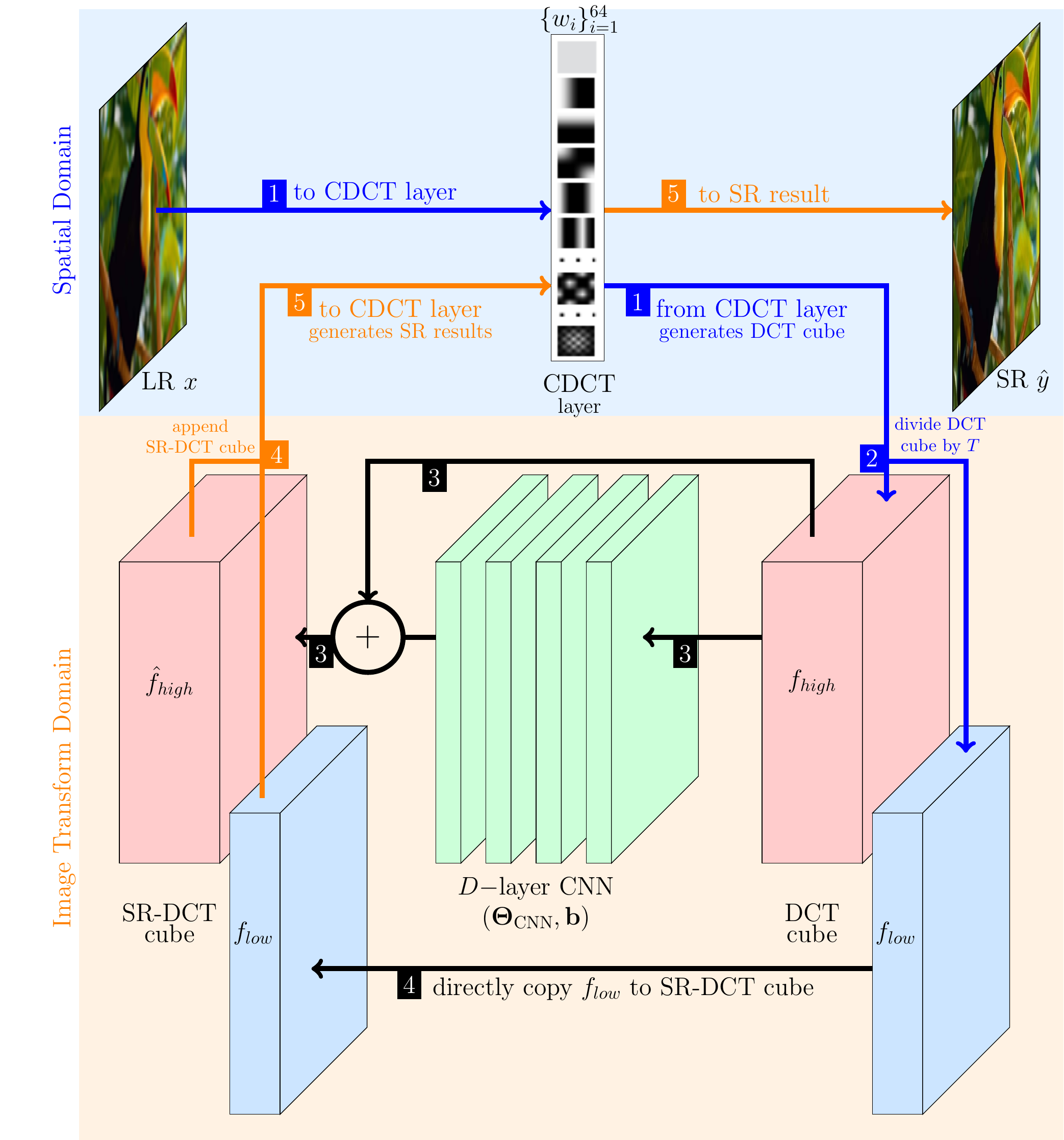}\sqz\sqz
\caption{\ninept The ORDSR network structure. Please refer to color version to follow the network flow. The CDCT layer serves two purposes: producing DCT cube (following blue arrow) and generating SR image from a SR-DCT cube (following orange arrow).}\label{fig:ORDSR}\sqz\sqz\sqz\sqz
\end{figure}

Fig. \ref{fig:ORDSR} shows the structure of the ORDSR network with $N=8$.
For an input LR image $x$, the goal of ORDSR is to generate its SR version $\hat{y}$ as follows: 
\begin{enumerate} \setlength{\parskip}{-5pt}\sqz
\item The input LR image $x$ is convolved with CDCT layer producing DCT cube $\{f_i\}_{i= 1}^{64}$ as shown in (\ref{equ:CDCTconv});
\item The DCT cube of $x$ is divided as $f_{low}$ and $f_{high}$ corresponding to low and high-frequency spectra by an index threshold $T$, based on description in Section \ref{sec:dctLayer};
\item The CNN takes $f_{high}$ as input and recovers the missing high-frequency information using a residual network structure, generating $\hat{f}_{high}$;
\item The $\hat{f}_{high}$ is appended by $f_{low}$ forming the SR-DCT cube $\{\hat{f}_i\}_{i= 1}^{64}$. As the $f_{low}$ is unchanged between $x$ and its corresponding HR image $y$, only $f_{high}$ needs to be modified for generating $\hat{y}$;
\item The SR-DCT cube $\{\hat{f}_i\}_{i= 1}^{64}$ is transpose convolved with CDCT layer (to perform the IDCT) generating $\hat{y}$, as shown in (\ref{equ:CDCTdeconv}).\sqz\sqz
\end{enumerate} 

In step 3, only taking the $f_{high}$ components of the DCT cube reduces the input channel numbers for the CNN, which makes the training procedure faster. In step 4, the CNN uses a residual network structure to further reduce the computational burden. Steps 1 and 5 are performed in the image spatial domain while steps 2-4 are in the image transform domain.

The inference procedure is denoted as $h_{(\mathbf{\Theta},\mathbf{b})}(x) = \hat{y}$, where $(\mathbf{\Theta},\mathbf{b})$ is the collection of all the trainable filter weights and biases of ORDSR network. Note that $\mathbf{\Theta} = \{\mathbf{\Theta}_{\text{CNN}}, \{w_i\}_{i=1}^{64}\}$ as shown in Fig. \ref{fig:ORDSR}.

We develop a modified back-propagation scheme \cite{techReport} which enables the proposed ORDSR to minimize: 
\begin{equation}
\label{eq:finalCostFn}
\resizebox{0.75\linewidth}{!}{$
\begin{split}
\mathbf{\Theta},\mathbf{b} = \underset{\mathbf{\Theta},\mathbf{b}}{argmin}\frac{1}{2}\|h_{(\mathbf{\Theta},\mathbf{b})}(x) - y\|^2_2 + \sigma\sum_{t}^{}\|\Theta_t\|^2_2\\\sqz\sqz\sqz\sqz+\gamma \sum_{(i,j)} \|vec(w_i)^Tvec(w_j)-\epsilon\|^2_2
\sqz\sqz\sqz\sqz\end{split}\sqz\sqz\sqz
$}
\end{equation}
where $\{(i,j)\}$ are all the unique pairwise indexes of the $\{w_i\}$ and $t$ is the collective index of all trainable filter weights in $\mathbf{\Theta}$. ORDSR also utilizes an $\ell_2$ regularization of weights with a trade off parameter $\sigma$. Note $w_i\in\mathbf{\Theta}$, thus filters of CDCT layer are updated to generate a better $\hat{y}$. 
\sqz\sqz 
\section{Experimental Results}
\sqz\sqz
\label{sec:exp}

\textbf{Data preparation}: The 291 images dataset \cite{Schulter_2015_CVPR} is used for training. The images are augmented by rotating the images by $\{90^{\circ}, 180^{\circ}, 270^{\circ}\}$ and scaling by factors of $\{0.7, 0.8, 0.9\}$. The augmented images are down-sampled and subsequently enlarged using bicubic interpolation to form the LR training images. 
All the LR/HR images are further cropped into $40\times40$ pixels sub-images with 10 pixels overlap for training. During the test phase, Set5 \cite{bevilacqua2012low} and Set14 \cite{zeyde2010single} are used to evaluate our proposed method. Both training and testing phases of ORDSR only utilize the luminance channel information. For color images, Cb, Cr channels are enlarged by bicubic interpolation. 

\textbf{Training Settings}:
During the training process, the gradients are clipped to 0.01 and the Adam optimizer \cite{kingma2014adam} is adopted to update $(\mathbf{\Theta}, \mathbf{b})$. The initial learning rate is 0.001 and decreases by $25\%$ every 25 epochs. $\sigma$ is set to $1 \times 10^{-3}$ to prevent over-fitting. The CNN has $D = 14$ same-sized convolutional hidden layers with filter size of $3 \times 3 \times 64$. This configuration results in a network with only 
$75\%$ of the parameters in the state-of-the-art method VDSR \cite{Kim_2016_VDSR}.
The ORDSR is implemented with TensorFlow \cite{tensorflow2015-whitepaper} packages on one TITAN X GPU for both the training and testing, which takes 5 hours to reach 85 epochs for the reported results.

\begin{figure}[H]\sqz\sqz\sqz\sqz
\centering
\includegraphics[trim=0cm 0cm 0cm 0cm,clip,width=\linewidth]{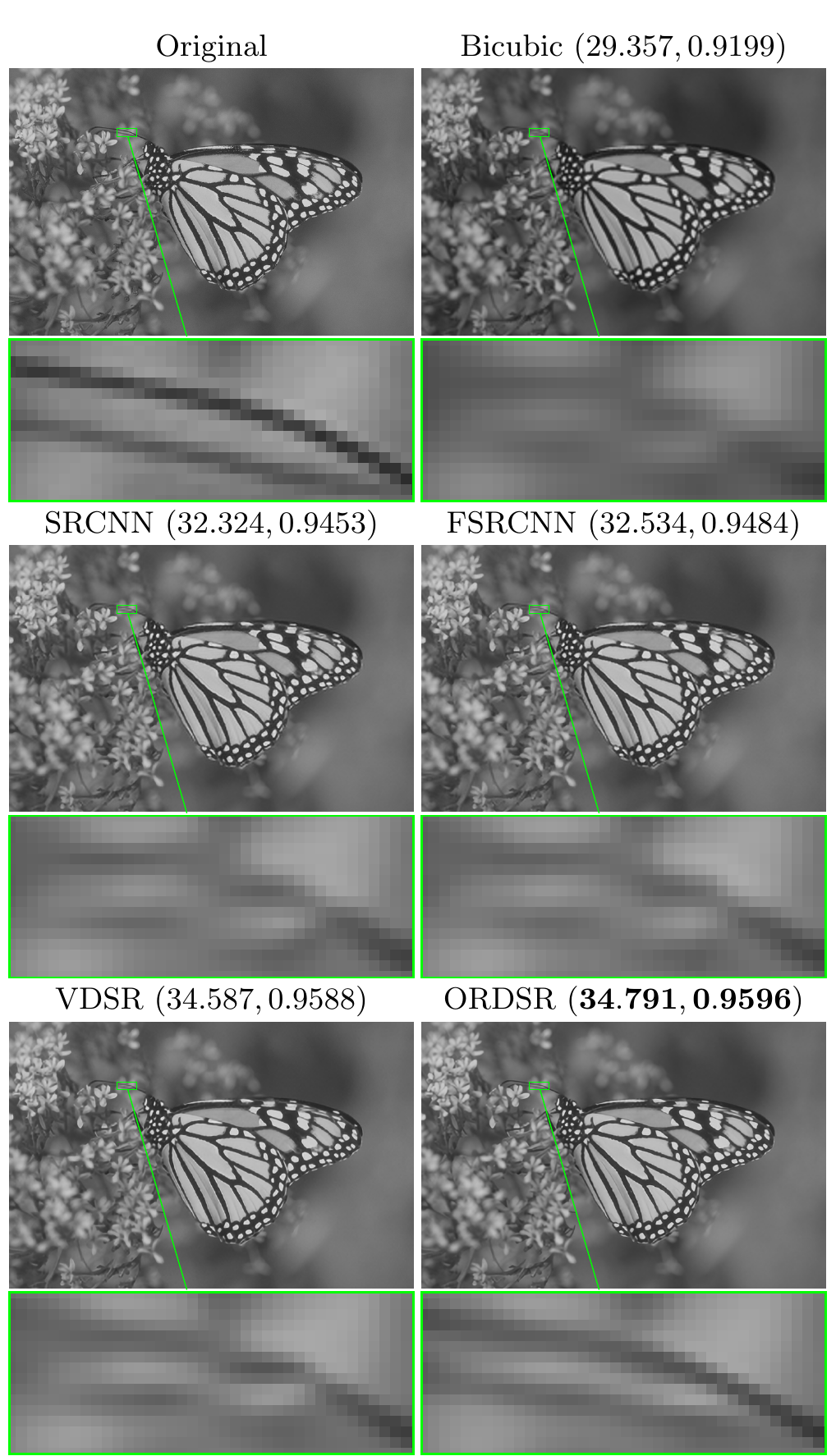}\sqz
\caption{\ninept The SR results of test image \textit{monarch.bmp} of scale factor 3. The image metrics is shown as (PSNR, SSIM). ORDSR produces best visual results with better quality assessments. }\label{fig:SR}\sqz\sqz\sqz
\end{figure}

\textbf{SR Results}:\footnote{\sqz\sqz\sqz Code available on \url{http://signal.ee.psu.edu/ORDSR.html}} Table \ref{tb:PSNR} shows the comparison of ORDSR with other state-of-the-art methods: classical methods ScSR \cite{yang2010image} and A+ \cite{timofte2014a+}, deep learning based methods SelfEx \cite{huang2015single}, FSRCNN \cite{dong2016accelerating}, SRCNN \cite{dong2014learning} and VDSR. The metrics used for image quality assessment are PSNR and SSIM \cite{wang2004image}. The comparison is constrained among methods with the same training set and same computational requirements. ORDSR produces best results using only $75\%$ the number of the parameters than VDSR. Fig. \ref{fig:SR} displays the testing image in detail, ORDSR generates more defined edges with better quality assessments among the competing methods. 

\begin{figure}\sqz\sqz\sqz
    \centering
    \begin{subfigure}{0.32\linewidth}
    \centering
    \includegraphics[width=\linewidth]{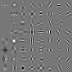}      
    \vspace{-.6cm}\caption{\ninept $\epsilon=0.0001$}
    \end{subfigure}
    \begin{subfigure}{0.32\linewidth}
    \centering
    \includegraphics[width=\linewidth]{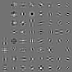}      
    \vspace{-.6cm}\caption{\ninept $\epsilon=0.001$}
    \end{subfigure}
    \begin{subfigure}{0.32\linewidth}
    \centering
    \includegraphics[width=\linewidth]{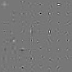}      
    \vspace{-.6cm}\caption{\ninept $\epsilon=0.01$}
    \end{subfigure}\sqz\sqz\sqz
    \caption{\ninept Learned filters of CDCT layers with different $\epsilon$. (Filters are normalized and reordered for display purpose.)}\label{fig:alpha}
    \sqz\sqz\sqz\sqz\sqz\sqz
\end{figure}

\textbf{CDCT Layer}: With different $\epsilon$, the orthogonality constants have different effects on the CDCT layer. As shown in Table \ref{tb:alpha}, with $\gamma=1$, a very small or very big $\epsilon$  will end up with a tightly constrained or relaxed CDCT layer. Fig. \ref{fig:alpha} shows smaller $\epsilon$ preserves more DCT filters structure within the CDCT layer. Cross validation shows $\epsilon = 0.001$ produces the best results. With $\gamma=0$, the ORDSR is trained without the orthogonality constraints, which produced less favorable results showing the importance of CDCT layer being orthogonal. Also if we excludes $w_i$ from $\mathbf{\Theta}$, the ORDSR is trained without updating the CDCT layer at all. Table \ref{tb:alpha} shows the importance of CDCT layer being adaptively trainable. 
\begin{table*}[t!]
\centering
\caption{\ninept PSNR and SSIM comparisons. (The best results are shown in bold \rd{\bf red} and the second best  are shown in \bl{blue}.)}\sqz\sqz\sqz
\label{tb:PSNR}\resizebox{\textwidth}{!}{
\begin{tabular}{|r|c||c||c||c||c||c||c||c||c|}
\hline
\begin{tabular}[c]{@{}c|c@{}}PSNR&SSIM\end{tabular}&Scale
                               & \begin{tabular}[c]{@{}c@{}}Bicubic\\{[}Baseline{]}\end{tabular}
                               & \begin{tabular}[c]{@{}c@{}}ScSR\\{[}TIP 10{]}\end{tabular}
                               & \begin{tabular}[c]{@{}c@{}}A+\\{[}ACCV 14{]}\end{tabular}
                               & \begin{tabular}[c]{@{}c@{}}SelfEx\\{[}CVPR 15{]}\end{tabular}
                               & \begin{tabular}[c]{@{}c@{}}FSRCNN\\{[}ECCV 16{]}\end{tabular}

                               & \begin{tabular}[c]{@{}c@{}}SRCNN\\{[}PAMI 16{]}\end{tabular}
                               & \begin{tabular}[c]{@{}c@{}}VDSR\\{[}CVPR 16{]}\end{tabular}
                               & \begin{tabular}[c]{@{}c@{}}ORDSR\\{[}Proposed{]}\end{tabular}                                   \\ \hline
Set5
& \begin{tabular}[c]{@{}c@{}}x2\\ x3\\ x4\end{tabular}
& \begin{tabular}[c]{@{}c|c@{}}33.64&0.9292\\ 30.39&0.8678\\ 28.42&0.8101\end{tabular}
& \begin{tabular}[c]{@{}c|c@{}}35.78&0.9485\\ 31.34&0.8869\\ 29.07&0.8263\end{tabular}
& \begin{tabular}[c]{@{}c|c@{}}36.55&0.9544\\ 32.58&0.9088\\ 30.27&0.8605\end{tabular}
& \begin{tabular}[c]{@{}c|c@{}}36.47&0.9538\\ 32.57&0.9092\\ 30.32&0.8640\end{tabular}
& \begin{tabular}[c]{@{}c|c@{}}36.94&0.9558\\ 33.06&0.9140\\ 30.55&0.8657\end{tabular} 
& \begin{tabular}[c]{@{}c|c@{}}36.66&0.9542\\ 32.75&0.9090\\ 30.48&0.8628\end{tabular}
& \begin{tabular}[c]{@{}c|c@{}}\bl{37.52}&\rd{\bf0.9586}\\\bl{33.66}&\bl{0.9212}\\ \bl{31.35}&\bl{0.8820}\end{tabular}
& \begin{tabular}[c]{@{}c|c@{}}\rd{\bf37.53}&\bl{0.9574}\\\rd{\bf 33.74}&\rd{\bf 0.9221}\\ \rd{\bf 31.45}&\rd{\bf 0.8847}\end{tabular} \\ \hline
Set14
& \begin{tabular}[c]{@{}c@{}}x2\\ x3\\ x4\end{tabular}
& \begin{tabular}[c]{@{}c|c@{}}30.22&0.8683\\ 27.53&0.7737\\ 25.99&0.7023\end{tabular}
& \begin{tabular}[c]{@{}c|c@{}}31.64&0.8940\\ 28.19&0.7977\\ 26.40&0.7218\end{tabular}
& \begin{tabular}[c]{@{}c|c@{}}32.29&0.9055\\ 29.13&0.8188\\ 27.33&0.7489\end{tabular}
& \begin{tabular}[c]{@{}c|c@{}}32.24&0.9032\\ 29.16&0.8196\\ 27.40&0.7518\end{tabular}
& \begin{tabular}[c]{@{}c|c@{}}32.54&0.9088\\ 29.37&0.8242\\ 27.50&0.7535\end{tabular} 
& \begin{tabular}[c]{@{}c|c@{}}32.42&0.9063\\ 29.28&0.8209\\ 27.40&0.7503\end{tabular}
& \begin{tabular}[c]{@{}c|c@{}}\bl{33.02}&\bl{0.9102}\\ \bl{29.75}&\bl{0.8294}\\ \bl{28.01}&\bl{0.7662}\end{tabular}
& \begin{tabular}[c]{@{}c|c@{}}\rd{\bf 33.04}&\rd{\bf 0.9109}\\ \rd{\bf 29.81}&\rd{\bf 0.8300}\\ \rd{\bf 28.06}&\rd{\bf 0.7664}\end{tabular} \\ \hline
\end{tabular}}
\end{table*}
\begin{table}[]\sqz
\sqz\sqz\sqz\sqz\sqz
\caption{\ninept Average results on Set14 with scale factor 3}\sqz\sqz\sqz
\label{tb:alpha}\resizebox{\linewidth}{!}{\begin{tabular}{|r|c|c|c|c|c|c||c||c|}\hline
 & \multicolumn{6}{c||}{$\gamma=1$} & \multirow{2}{*}{$\gamma=0$} & \multirow{2}{*}{$w_i\notin\mathbf{\Theta}$}\\  \cline{2-7}
 $\epsilon=$&0.0001&\bf 0.001& 0.01& 0.1& 0.5& 1& &\\\hline
PSNR&29.7932&\bf 29.8104& 29.7815& 29.7805&29.7786&29.7208&29.7621&29.7165\\\hline
SSIM&0.8295&\bf 0.8300& 0.8281& 0.8265&0.8266&0.8252&0.8201&0.8189\\\hline
\end{tabular}}
\end{table}

\begin{figure}[t]
\sqz\sqz\sqz\sqz
\centering
\includegraphics[trim=0cm 0cm 2cm 0cm,clip,width=\linewidth]{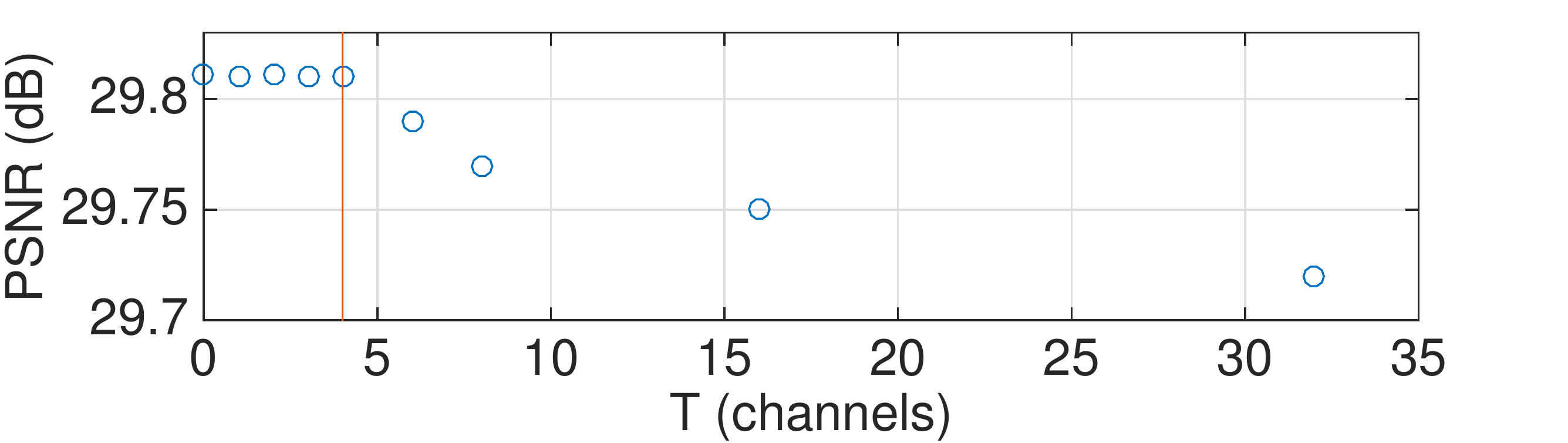}\sqz\sqz
\caption{\ninept Avg. PSNR of Set14 with scale factor 3 on different $T$. When $T<4$ , decreasing $T$ will not affect the SR results.}\label{fig:T}\sqz\sqz\sqz\sqz
\end{figure}
\noindent \textbf{Threshold $T$}: Fig. \ref{fig:T} shows the effects of $T$ over the PSNR of the SR results. A smaller $T$ implies a smaller fraction of $f_{low}$ is directly copied to SR-DCT cube as described in the inference step 3. However, after $T<4$, decreasing the threshold does not change the SR image quality significantly. This further shows that the low frequency spectra between LR/HR image are indeed shared. All reported results use $T=4$. \sqz\sqz\sqz\sqz\sqz
\section{Conclusion}\sqz\sqz\sqz
We propose a novel network structure to tackle SR problem in the image transform domain. We show that DCT can be integrated into the network structure as a Convolutional DCT (CDCT) layer. We further extend the network to allow the CDCT layer to become {\em trainable (i.e. optimizable)}. Experimental results show the effectiveness of performing SR in the image transform domain by ORDSR, also the significance of ORDSR learning bases that are specific for natural image SR.

\vfill\pagebreak



\ninept
\bibliographystyle{IEEEbib_ini}
\bibliography{refs.bib}

\begin{thebibliography}{10}

\bibitem{park2003super}
S.~C. Park, M.~K. Park, and M.~G. Kang,
\newblock ``Super-resolution image reconstruction: a technical overview,''
\newblock {\em Signal Processing Magazine, IEEE}, vol. 20, no. 3, pp. 21--36,
  2003.

\bibitem{mallat2010super}
S. Mallat and G. Yu,
\newblock ``Super-resolution with sparse mixing estimators,''
\newblock {\em Image Processing, IEEE Transactions on}, vol. 19, no. 11, pp.
  2889--2900, 2010.

\bibitem{chang2004super}
H. Chang, D.-Y. Yeung, and Y. Xiong,
\newblock ``Super-resolution through neighbor embedding,''
\newblock in {\em Computer Vision and Pattern Recognition, IEEE Conference on},
  2004, vol.~1, pp. I--I.

\bibitem{glasner2009super}
D. Glasner, S. Bagon, and M. Irani,
\newblock ``Super-resolution from a single image,''
\newblock in {\em Computer Vision, IEEE International Conference on}, 2009, pp.
  349--356.

\bibitem{yang2010image}
J. Yang, J. Wright, T.~S. Huang, and Y. Ma,
\newblock ``Image super-resolution via sparse representation,''
\newblock {\em Image Processing, IEEE Transactions on}, vol. 19, no. 11, pp.
  2861--2873, 2010.

\bibitem{kim2010single}
K.~I. Kim and Y. Kwon,
\newblock ``Single-image super-resolution using sparse regression and natural
  image prior,''
\newblock {\em Pattern Analysis and Machine Intelligence, IEEE Transactions
  on}, vol. 32, no. 6, pp. 1127--1133, 2010.

\bibitem{zhang17image}
L. Zhang and W. Zuo,
\newblock ``Image restoration: From sparse and low-rank priors to deep priors,
  lecture notes,''
\newblock {\em Signal Processing Magazine, IEEE}, vol. 34, no. 5, pp. 172--179,
  2017.

\bibitem{Timofte_2017_CVPR_Workshops}
R. Timofte, E. Agustsson, L. Van~Gool, M.-H. Yang, L. Zhang, et~al.,
\newblock ``Ntire 2017 challenge on single image super-resolution: Methods and
  results,''
\newblock in {\em Computer Vision and Pattern Recognition Workshops, IEEE
  Conference on}, July 2017.

\bibitem{dong2014learning}
C. Dong, C.~C. Loy, K. He, and X. Tang,
\newblock ``Learning a deep convolutional network for image super-resolution,''
\newblock in {\em Computer Vision, ECCV}, pp. 184--199. Springer, 2014.

\bibitem{dong2016accelerating}
C. Dong, C.~C. Loy, and X. Tang,
\newblock ``Accelerating the super-resolution convolutional neural network,''
\newblock in {\em Computer Vision, ECCV}, pp. 391--407. 2016.

\bibitem{tiantong16deep}
T. Guo, H.~S. Mousavi, and V. Monga,
\newblock ``Deep learning based image super-resolution with coupled
  backpropagation,''
\newblock in {\em Signal and Information Processing, IEEE Global Conference
  on}, 2016, pp. 237--241.

\bibitem{huang2015single}
J.-B. Huang, A. Singh, and N. Ahuja,
\newblock ``Single image super-resolution from transformed self-exemplars,''
\newblock in {\em Computer Vision and Pattern Recognition, IEEE Conference on},
  2015, pp. 5197--5206.

\bibitem{wang2015self}
Z. Wang, Y. Yang, Z. Wang, S. Chang, W. Han, J. Yang, and T.~S. Huang,
\newblock ``Self-tuned deep super resolution,''
\newblock {\em arXiv preprint arXiv:1504.05632}, 2015.

\bibitem{Kim_2016_VDSR}
J. Kim, J.~K. Lee, and K.~M. Lee,
\newblock ``Accurate image super-resolution using very deep convolutional
  networks,''
\newblock in {\em Computer Vision and Pattern Recognition, IEEE Conference on},
  2016, pp. 1646--1654.

\bibitem{zhao2003wavelet}
S. Zhao, H. Han, and S. Peng,
\newblock ``Wavelet-domain hmt-based image super-resolution,''
\newblock in {\em Image Processing, IEEE International Conference on}, 2003,
  pp. II--953.

\bibitem{robinson2010efficient}
M.~D. Robinson, C.~A. Toth, J.~Y. Lo, and S. Farsiu,
\newblock ``Efficient fourier-wavelet super-resolution,''
\newblock {\em Image Processing, IEEE Transactions on}, vol. 19, no. 10, pp.
  2669--2681, 2010.

\bibitem{wahed2007image}
M.~E.-S. Wahed,
\newblock ``Image enhancement using second generation wavelet super
  resolution,''
\newblock {\em International Journal of Physical Sciences}, vol. 2, no. 6, pp.
  149--158, 2007.

\bibitem{ji2009robust}
H. Ji and C. Ferm{\"u}ller,
\newblock ``Robust wavelet-based super-resolution reconstruction: theory and
  algorithm,''
\newblock {\em Pattern Analysis and Machine Intelligence, IEEE Transactions
  on}, vol. 31, no. 4, pp. 649--660, 2009.

\bibitem{guo2017deep}
T. Guo, H.~S. Mousavi, T.~H. Vu, and V. Monga,
\newblock ``Deep wavelet prediction for image super-resolution,''
\newblock in {\em Computer Vision and Pattern Recognition Workshops, IEEE
  Conference on}, 2017, pp. 1100--1109.

\bibitem{noh2015learning}
H. Noh, S. Hong, and B. Han,
\newblock ``Learning deconvolution network for semantic segmentation,''
\newblock in {\em Computer Vision, IEEE International Conference on}, 2015, pp.
  1520--1528.

\bibitem{dumoulin2016guide}
V. Dumoulin and F. Visin,
\newblock ``A guide to convolution arithmetic for deep learning,''
\newblock {\em arXiv preprint arXiv:1603.07285}, 2016.

\bibitem{techReport}
T. Guo, H.~S. Mousavi, and V. Monga,
\newblock ``A technical report on: orthogonally regularized deep networks for
  image super-resolution,'' 2017,
\newblock Code available on http://signal.ee.psu.edu/ORDSR.html.

\bibitem{Schulter_2015_CVPR}
S. Schulter, C. Leistner, and H. Bischof,
\newblock ``Fast and accurate image upscaling with super-resolution forests,''
\newblock in {\em Computer Vision and Pattern Recognition, IEEE Conference on},
  2015, pp. 3791--3799.

\bibitem{bevilacqua2012low}
M. Bevilacqua, A. Roumy, C. Guillemot, and M.~L. Alberi-Morel,
\newblock ``Low-complexity single-image super-resolution based on nonnegative
  neighbor embedding,''
\newblock 2012.

\bibitem{zeyde2010single}
R. Zeyde, M. Elad, and M. Protter,
\newblock ``On single image scale-up using sparse-representations,''
\newblock in {\em International conference on curves and surfaces}. Springer,
  2010, pp. 711--730.

\bibitem{kingma2014adam}
D. Kingma and J. Ba,
\newblock ``Adam: A method for stochastic optimization,''
\newblock {\em arXiv preprint arXiv:1412.6980}, 2014.

\bibitem{tensorflow2015-whitepaper}
M. Abadi, A. Agarwal, and P.~B. et. al.,
\newblock ``{TensorFlow}: Large-scale machine learning on heterogeneous
  systems,'' 2015,
\newblock Software available from tensorflow.org.

\bibitem{timofte2014a+}
R. Timofte, V. De~Smet, and L. Van~Gool,
\newblock ``A+: Adjusted anchored neighborhood regression for fast
  super-resolution,''
\newblock in {\em Computer Vision, ACCV}, pp. 111--126. Springer, 2014.

\bibitem{wang2004image}
Z. Wang, A.~C. Bovik, H.~R. Sheikh, and E.~P. Simoncelli,
\newblock ``Image quality assessment: from error visibility to structural
  similarity,''
\newblock {\em Image Processing, IEEE Transactions on}, vol. 13, no. 4, pp.
  600--612, 2004.

\end{thebibliography}

\end{document}